%% file: main.tex
\documentclass[journal]{IEEEtran}
\usepackage[utf8]{inputenc}
\usepackage{amsmath,amssymb,graphicx}
\usepackage{nicefrac}
\usepackage{mathtools}
\usepackage{caption}
\usepackage[font=footnotesize]{subcaption}
\usepackage{soul}
\usepackage{mwe}
\usepackage{booktabs}
\usepackage{color}
\usepackage{setspace}
\usepackage[inline]{enumitem}
\usepackage{hyperref}
\usepackage{multirow}
\definecolor{darkgreen}{rgb}{0,0.6,0.2}

\usepackage{algorithm}
\usepackage{algorithmic}
\usepackage{array}
\usepackage{tikz}
\usepackage{pgfplots}
\pgfplotsset{compat=1.14}
\usetikzlibrary{calc}

\usepackage{alphalph}

\usepackage{url}

\title{Code-Aligned Autoencoders for Unsupervised Change Detection in Multimodal Remote Sensing Images}

\author{Luigi~T.~Luppino,~
        Mads~A.~Hansen,~
        Michael~Kampffmeyer,\\
        Filippo~M.~Bianchi,~
        Gabriele~Moser,~
        Robert~Jenssen,~
        and~Stian~Normann~Anfinsen
\thanks{L.T.~Luppino, M.A.~Hansen, M.~Kampffmeyer, R.~Jenssen and S.N.~Anfinsen are with the Machine Learning Group, Department of Physics and Technology, UiT The Arctic University of Norway, e-mail: luigi.t.luppino@uit.no.}
\thanks{F.M.~Bianchi is with NORCE Norwegian Research Center, Norway.}%
\thanks{G.~Moser is with DITEN Department, University of Genoa, Italy.}
\thanks{Manuscript received -; revised -.}}

\begin{document}

\maketitle

\begin{abstract}
Image translation with convolutional autoencoders has recently been used as an approach to multimodal change detection in bitemporal satellite images.
A main challenge is the alignment of the code spaces by reducing the contribution of change pixels to the learning of the translation function.
Many existing approaches train the networks by exploiting supervised information of the change areas, which, however, is not always available.
We propose to extract relational pixel information captured by domain-specific affinity matrices at the input and use this to enforce alignment of the code spaces and reduce the impact of change pixels on the learning objective. A change prior is derived in an unsupervised fashion from pixel pair affinities that are comparable across domains.
To achieve code space alignment we enforce that pixel with similar affinity relations in the input domains should be correlated also in code space. We demonstrate the utility of this procedure in combination with cycle consistency.
The proposed approach are compared with state-of-the-art deep learning algorithms.
Experiments conducted on four real datasets show the effectiveness of our methodology.
\end{abstract}


\input{Introduction.tex}

\input{Methodology.tex}

\input{Results.tex}

\input{Conclusions.tex}

\bibliographystyle{IEEEtran}
\bibliography{references}
\end{document}

%% file: Introduction.tex
\section{Introduction}\label{sec:intro}

\IEEEPARstart{C}{hange} detection (CD) methods in remote sensing aim at identifying changes happening on the Earth by comparing two or more images acquired at different times~\cite{mercier2008conditional}.
Multitemporal analyses with satellite data include land use mapping of urban and agricultural areas~\cite{li2004analyzing,herold2002use}, and monitoring of large scale changes such as deforestation~\cite{khan2017forest}, lake and glacier reduction~\cite{alesheikh2007coastline, berthier2007remote}, urbanisation~\cite{griffiths2010mapping}, etc.
Bitemporal applications mainly concerned with the detection and assessment of natural disasters and sudden events, like earthquakes~\cite{brunner2010earthquake}, floods~\cite{luppino2019unsupervised}, forest fires~\cite{volpi2015spectral}, and so forth.

Traditional CD methods rely on homogeneous data, namely a set of images acquired by the same sensor, under the same geometry, seasonal conditions, and recording settings.
However, these restrictions are too strong for many practical examples.
First of all, the satellite revisit period sets the upper limit to the temporal resolution when monitoring long-term trends, and the lower limit to the response time when assessing the damages of sudden events.
Moreover, even when two images are collected with the same configurations, they might be not homogeneous because of other factors, for example light conditions for optical data or humidity and precipitation for synthetic aperture radar (SAR).

Heterogeneous CD methods overcome these limitations, but at the cost of having to handle more complicated issues;
Heterogeneous data imply different domains, diverse statistical distributions and inconsistent class signatures across the two images, especially when different sensors are involved, which makes a direct comparison infeasible~\cite{luppino2018remote}.
These problems have been tackled by use of many different techniques: copula theory~\cite{mercier2008conditional}, marginal densities transformations~\cite{storvik2009combination}, evidence theory~\cite{liu2011dynamic,liu2014change}, graph theory~\cite{tuia2013graph}, manifold learning~\cite{prendes2015new}, kernelised or deep canonical correlation analysis~\cite{volpi2015spectral,zhou2019cross,yang2018heterogeneous}, dictionary learning~\cite{gong2016coupled}, scale-invariant local descriptors~\cite{liu2016unsupervised,liu2019contrario}, superpixel segmentation~\cite{marcos2016geospatial}, clustering~\cite{luppino2017clustering}, minimum energy~\cite{touati2018energy}, multidimensional scaling~\cite{touati2018change}, nonlinear regression~\cite{liu2018change,luppino2019unsupervised}, and deep learning (especially autoencoders)~\cite{zhang2016cd,liu2016deep,zhao2017discriminative,su2017deep,zhan2018log,zhan2018iterative}.

A common solution in heterogeneous CD is to apply highly nonlinear transformations to transfer the data from one domain to the other and vice versa~\cite{su2017deep,niu2018conditional,gong2019coupling}.
Alternatively, all the data are mapped to a common domain where they can be compared~\cite{storvik2009combination,zhang2016cd,liu2016deep,zhan2018iterative}.
Nonetheless, this crucial step often requires iterative fine-tuning of the transformation functions starting from unreliable preliminary results, e.g.\ random initialisation~\cite{liu2016deep,zhan2018iterative} and clustering~\cite{su2017deep}, or from manually selected training samples~\cite{mercier2008conditional,volpi2015spectral,prendes2015new} that are not always available.

One contemporary way to map data across two domains is image-to-image (I2I) translation using a conditional generative adversarial network (cGAN)~\cite{Isola_2017_CVPR}, which was extended by enforcing cyclic consistency in the cycleGAN architecture ~\cite{zhu2017unpaired}. These approaches have inspired many recent heterogeneous CD methods~\cite{niu2018conditional,gong2019coupling,luppino2020deep}. 
A notable difference between the cGAN and the cycleGAN is that training of the former requires paired images that contain the same objects imaged with different styles or sensor modes,
whereas the cycleGAN does not. Paired I2I translation can only be applied in heterogeneous CD if change pixels are censored, as these will otherwise distort the training process and promote a transformation between different objects.

When generative adversarial frameworks are used in heterogeneous CD, the translated (or cyclically translated) images take the role as fake or generated data, and the network is trained to make them indistinguishable from true images from the relevant domain. The cGAN and cycleGAN may succeed to align the distributions of translated data and true data, but they are also seen to suffer from inherent drawbacks: They rely on large training sets, the iterative training of generator and discriminator must be judiciously balanced, training is prone to mode collapse, and reasonable values of the hyperparameters can be difficult to find due to oscillating and unstable behaviour of the loss function. We therefore seek alternative training strategies to the adversarial ones.

In this work, we propose a simple unsupervised, heterogeneous CD method, inspired by the paradigm of I2I translation.
The idea is to align the code layers of two autoencoders and treat them as a common latent space, so that the output of one encoder can be the input of both decoders, leading in one case to reconstruction of data in their original domain, and in the other case to their transformation into the other domain.
Local information extracted directly from the input images is exploited to drive the code alignment in an unsupervised manner.
Specifically, affinity matrices of the training patches are computed and compared, and the extracted information is used to ensure that pixel pairs that are similar in both input domains also have a high correlation in the common latent space. The implementation of this principle is inspired by the deep kernelised autoencoder of Kampffmeyer et al.~\cite{kampffmeyer2018deep, BIANCHI2019106973}, where the inner product between the codes produced by two datapoints is forced to match their precomputed affinity.

To summarise, the contributions of this work are the following:

\begin{itemize}
    \item We propose a simple, yet effective loss term, able to align the latent spaces of two autoencoders in an unsupervised manner.
    \item We implement a deep neural network for heterogeneous CD that incorporates this loss term.
    \item The well-documented TensorFlow 2.0 framework that we provide can be easily used for the development of other CD methods and for direct comparison with ours. Source code is made available at \url{https://github.com/llu025/Heterogeneous_CD}.
\end{itemize}

The remainder of this paper is organised as follows: The core ideas and the main contribution are presented in Sec.\ \ref{sec:meth};
Experiments were conducted on four different real datasets, and Sec.\ \ref{sec:results} shows the results of the proposed approach against several state-of-the-art methods;
Sec.\ \ref{sec:conclusions} concludes the paper.

%% file: Methodology.tex
\section{Methodology}\label{sec:meth}

\begin{figure*}[ht!]
\begin{subfigure}[t]{0.45\linewidth}
\includegraphics[width=\linewidth,keepaspectratio]{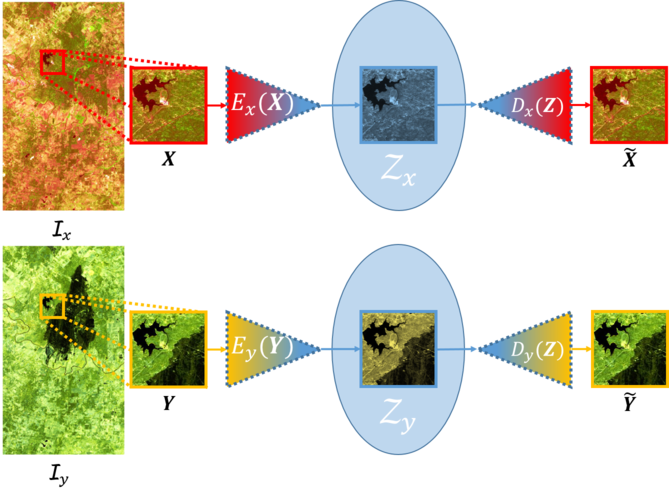}
\caption{Not aligned code spaces}
\label{fig:issue}
\end{subfigure}
\hfill
\begin{subfigure}[t]{0.45\linewidth}
\includegraphics[width=\linewidth,height=0.73\linewidth]{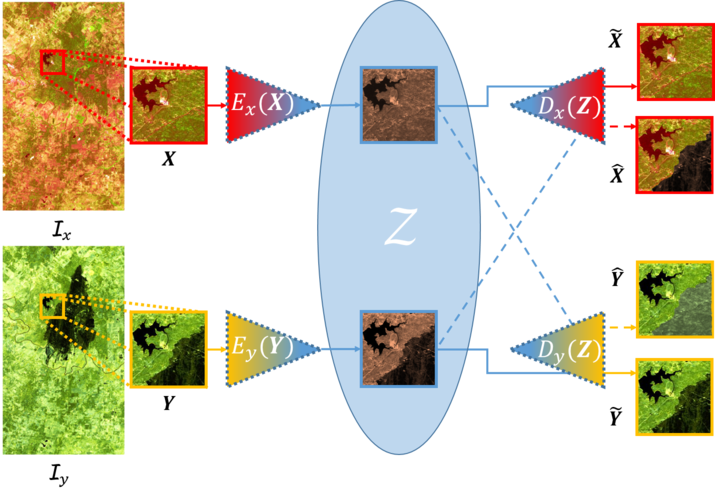}
\caption{Aligned code spaces}
\label{fig:goal}
\end{subfigure}
\caption{Two autoencoders without (a) and with (b) code space alignment.}
\label{fig:alignment}
\end{figure*}
Assume that we have two different sensors (or sensor modes) whose single-pixel measurements lie in the domains $\mathcal{X}$ and $\mathcal{Y}$. These could be e.g.\ $\mathbb{R}_{\geq0}$ (nonnegative real numbers) for a single-channel SAR sensor, $\mathbb{R}_{\geq0}^C$ for a multispectral radiometer with $C$ bands, or $\mathbb{C}_{\succeq0}^{C\times C}$ for a polarimetric SAR sensor with $C$ polarisations that records a complex and semipositive definite covariance matrix for each pixel.

Further assume that these sensors are scanning the same geographical region at separate times and we obtain an image $\mathcal{I_X}\in\mathcal{X}^{H\times W}$ recorded at time $t_1$ and an image $\mathcal{I_Y}\in\mathcal{Y}^{H\times W}$ recorded at $t_2\!>\!t_1$. The images and their domains have common dimensions, the shared height $H$ and width $W$, which are obtained after coregistration and resampling. They will in general have different numbers of channels, denoted as $|\mathcal{X}|$ and $|\mathcal{Y}|$.
The two images can be thought of as realisations of stochastic processes that generate data tensors from domain $\mathcal{X}$ and $\mathcal{Y}$.

An underlying assumption is that a limited part of the image has changed between $t_1$ and $t_2$.
The final goal is to detect all changes in the scene.
However, given the heterogeneity of $\mathcal{X}$ and $\mathcal{Y}$, direct comparison is meaningless, if not unfeasible, without any preprocessing step.
Let $\boldsymbol{X}\in\mathcal{X}^{h\times w}$ and $\boldsymbol{Y}\in\mathcal{Y}^{h\times w}$ be data tensors holding size $h\times w$ patches of the full images $\mathcal{I_X}$ and $\mathcal{I_Y}$.
We are interested in implementing the two transformations: $\boldsymbol{\hat{Y}} = F(\boldsymbol{X})$ and $\hat{\boldsymbol{X}} = G(\boldsymbol{Y})$, defined as $F:\mathcal{X}^{h\times w}\to\mathcal{Y}^{h\times w}$ and $G:\mathcal{Y}^{h\times w}\to\mathcal{X}^{h\times w}$, 
%
%
to map data between the image domains.
In this way, the input images can be transferred to the opposite domain, and the changes can be detected by computing the difference image as the weighted average:
%
\begin{equation}
    \boldsymbol{\Delta} = W_{\mathcal{X}} \cdot d^\mathcal{X}(\boldsymbol{X},\boldsymbol{\hat{X}}) + W_{\mathcal{Y}} \cdot d^\mathcal{Y}(\boldsymbol{Y},\boldsymbol{\hat{Y}})\,,
\end{equation}
where $d^\mathcal{X}(\cdot,\cdot)$ and $d^\mathcal{Y}(\cdot,\cdot)$ are sensor-specific distances, chosen according to the statistical distribution of the data, which operate pixel-wise. 
The generic weights $W_{\mathcal{X}}$ and $W_{\mathcal{X}}$ can be used to balance the contribution of the domain-specific distances. We may want to use $W_{\mathcal{X}}=1/|\mathcal{X}|$ and $W_{\mathcal{Y}}=1/|\mathcal{Y}|$ in order to remove undue influence of the number of channels if $d^\mathcal{X}$ and $d^\mathcal{Y}$ involve summations on the corresponding channels. Alternatively, it may be appropriate to compensate for different noise levels of the sensors that affect the magnitude of the distances, for instance by boosting the contribution of optical data with respect to highly speckled radar data. The weights can be set heuristically or according to empirical optimisation and theoretical considerations. We prefer to use $L_2$ distances to limit the computational cost.

To implement $F(\boldsymbol{X})$ and $G(\boldsymbol{Y})$, we use a framework that consists of two autoencoders, each associated with one of the two image domains $\mathcal{X}$ and $\mathcal{Y}$ (We will from now suppress the superscripting with image patch dimensions $h\times w$).
Specifically, they consist of two encoder-decoder pairs implemented as deep neural networks: the encoder $E_{\mathcal{X}}(\boldsymbol{X}):\mathcal{X}\to\mathcal{Z_X}$ and decoder $D_{\mathcal{X}}(\boldsymbol{Z}):\mathcal{Z_X}\to\mathcal{X}$; the encoder $E_{\mathcal{Y}}(\boldsymbol{Y}):\mathcal{Y}\to\mathcal{Z_Y}$ and decoder $D_{\mathcal{Y}}(\boldsymbol{Z}):\mathcal{Z_Y}\to\mathcal{X}$.
Here, $\mathcal{Z_X}$ and $\mathcal{Z_Y}$ denote the code layer or latent space domains of the respective autoencoders. These are implemented with common dimensions, such that the code layer representation $\boldsymbol{Z}$ (also known as the \emph{code}) can denote data tensors in both $\mathcal{Z_X}$ and $\mathcal{Z_Y}$. When we need to specify which input space the codes originate from, they will be written as $\boldsymbol{Z}^\mathcal{X}$ and $\boldsymbol{Z}^\mathcal{Y}$.

When trained separately and under the appropriate regularisation, the autoencoders will learn to encode their inputs and reconstruct them with high fidelity in output. Without any external forcing, the distributions of the codes in $\mathcal{Z_X}$ and $\mathcal{Z_Y}$ will in general not be close (see Fig.\ \ref{fig:issue} for a visual example). However, we will introduce loss terms that enforce their alignment, both in distribution and in the location of land covers within the distributions\footnote{Alignment in distribution is not sufficient, since the arrangement of land covers within the distributions may have changed, for instance by mode swapping.}. If the code distributions in $\mathcal{Z_X}$ and $\mathcal{Z_Y}$ align successfully, the encoders can be cascaded with the adjacent decoders to map the latent domain codes back to their original domains, or with the opposite decoders to map data across domains, leading to the sought transformations:
\begin{equation}
\begin{split}
    \boldsymbol{\hat{Y}} & = F(\boldsymbol{X}) = D_{\mathcal{Y}}(\boldsymbol{Z}^\mathcal{X}) = D_{\mathcal{Y}}\left(E_{\mathcal{X}}\left(\boldsymbol{X}\right)\right)\,, \\
    \boldsymbol{\hat{X}} & = G(\boldsymbol{Y}) = D_{\mathcal{X}}(\boldsymbol{Z}^\mathcal{Y}) = D_{\mathcal{X}}\left(E_{\mathcal{Y}}\left(\boldsymbol{Y}\right)\right)\,,
\end{split}
\end{equation}
as depicted in Fig.\ \ref{fig:goal}.

Autoencoders require regularisation in order to avoid learning an identity mapping. This is commonly implemented as sparsity constraints or compression at the code layer by dimensionality reduction, with the latter measure known as a bottleneck. In our implementation, we retain the image patch dimensions ($h$ and $w$) throughout the hidden layers of the autoencoder and do not resort to bottlenecking, as this is seen to produce the best results. The additional constraints associated with code alignment and crossdomain mapping are seen to enforce the required regularisation.

In the following, we define the terms of the loss function $\mathcal{L}\left(\boldsymbol{\vartheta}\right)$.
The loss function is minimised with respect to the parameters of the networks, $\boldsymbol{\vartheta}$, to train the two autoencoders with the goal of obtaining the desired $F(\boldsymbol{X})$ and $G(\boldsymbol{Y})$.
In order to compare input patches and translated ones, a weighted distance between patches is defined.
Let $\boldsymbol{A}$ and $\boldsymbol{B}$ be two equal-sized $h \times w$ patches, then $\delta(\boldsymbol{A},\boldsymbol{B}|\boldsymbol{\pi})$ denotes a general weighted distance between patches, where $\boldsymbol{\pi}$ is a vector of weights, each associated with a pixel $i\in\{1,\dots,n\}$ of the patches, with $n=h\cdot w$.
In particular, $\delta(\boldsymbol{A},\boldsymbol{B}|\boldsymbol{1})=\delta(\boldsymbol{A},\boldsymbol{B})$, being $\boldsymbol{1}$ a vector of ones.
When the pixel measurements $\boldsymbol{a}_i\in\boldsymbol{A}$ and $\boldsymbol{b}_i\in\boldsymbol{B}$ are vectors, the mean squared $L_2$ norm can be used:
\begin{equation}
\delta(\boldsymbol{A},\boldsymbol{B}|\boldsymbol{\pi}) = \frac{1}{n}\displaystyle\sum_{i=1}^{n}\pi_i\lVert\boldsymbol{a}_i-\boldsymbol{b}_i\rVert_2^2\,.
\end{equation}

\subsection{Reconstruction Loss}

Consider two training patches of $h \times w$ pixels extracted at the same location from $\mathcal{I_X}$ and $\mathcal{I_Y}$. The first requirement for the autoencoders is to reproduce their input as faithfully as possible in output, which means that for the reconstructed image patches $\boldsymbol{\tilde{X}}$ and $\boldsymbol{\tilde{Y}}$,
\begin{equation}
\begin{split}
    \boldsymbol{\tilde{\boldsymbol{X}}} = D_{\mathcal{X}}\left(E_{\mathcal{X}}\left(\boldsymbol{X}\right)\right) \simeq \boldsymbol{X} \\
    \boldsymbol{\tilde{Y}} = D_{\mathcal{Y}}\left(E_{\mathcal{Y}}\left(\boldsymbol{Y}\right)\right) \simeq \boldsymbol{Y}
\end{split}
\end{equation}
must hold true.
We introduce the mean squared error between the desired and the predicted output as the reconstruction loss term:
\begin{equation}\label{eq:recon}
    \mathcal{L}_{\mathrm r}(\boldsymbol{\vartheta}) = \mathbb{E}_{\boldsymbol{X}}\left[\delta( \tilde{\boldsymbol{X}},\boldsymbol{X})\right] +
    \mathbb{E}_{\boldsymbol{Y}}\left[\delta(\tilde{\boldsymbol{Y}},\boldsymbol{Y})\right].
\end{equation}

\subsection{Cycle-consistency Loss}

Cycle-consistency implies that data transformed from $\mathcal{X}$ to $\mathcal{Y}$ and back to $\mathcal{X}$ should match exactly the input data we started from. The same applies to the transformations from $\mathcal{Y}$ to $\mathcal{X}$ and back.
If $F(\boldsymbol{X})$ and $G(\boldsymbol{Y})$ are close to be perfectly adapted, it must hold true that
\begin{equation}
\begin{split}
    \dot{\boldsymbol{X}} = G(\boldsymbol{\hat{Y}}) = G\left(F(\boldsymbol{X})\right) \simeq \boldsymbol{X}\,, \\
    \dot{\boldsymbol{Y}} = F(\boldsymbol{\hat{X}}) = F\left(G(\boldsymbol{Y})\right) \simeq \boldsymbol{Y}\,,
\end{split}
\end{equation}
where $\dot{\boldsymbol{X}} = G(\hat{\boldsymbol{Y}})$ and $\dot{\boldsymbol{Y}} = F(\hat{\boldsymbol{X}})$ indicate the data cyclically transformed to the original domains.
Hence, we define the cycle-consistency loss term as:
\begin{equation}
    \mathcal{L}_{\mathrm c}(\boldsymbol{\vartheta}) = \mathbb{E}_{\boldsymbol{X}}\left[\delta(\dot{\boldsymbol{X}},\boldsymbol{X})\right] +
    \mathbb{E}_{\boldsymbol{Y}}\left[\delta( \dot{\boldsymbol{Y}},\boldsymbol{Y})\right]\,.
\end{equation}
We note that cycle-consistency, like reconstruction, can be evaluated with unpaired data, since $\tilde{\boldsymbol{X}}$ and $\dot{\boldsymbol{X}}$ are computed from $\boldsymbol{X}$ while $\tilde{\boldsymbol{Y}}$ and $\dot{\boldsymbol{Y}}$ are computed from $\boldsymbol{Y}$.

\subsection{Weighted Translation Loss}

For those pixels not affected by changes, we require
\begin{equation}
\begin{split}
    \hat{\boldsymbol{Y}} &= F(\boldsymbol{X}) \simeq \boldsymbol{Y} \\
    \hat{\boldsymbol{X}} &= G(\boldsymbol{Y}) \simeq \boldsymbol{X}\,.
\end{split}
\end{equation}
From the opposite perspective, pixels that are likely to be changed shall not fulfil these same requirements.
Thus, the weighted translation loss term is defined as follows:
\begin{equation}
    \mathcal{L}_{\mathrm t}(\boldsymbol{\vartheta}) = \mathbb{E}_{\boldsymbol{X},\boldsymbol{Y}}\left[\delta(\hat{\boldsymbol{X}},\boldsymbol{X}|\boldsymbol{\pi})\right] +
    \mathbb{E}_{\boldsymbol{X},\boldsymbol{Y}}\left[\delta(\hat{\boldsymbol{Y}},\boldsymbol{Y}|\boldsymbol{\pi})\right] \,,
\end{equation}
where the contribution to the translation loss of the pixels is weighted by the prior $\boldsymbol{\pi}$, whose elements $\{\pi_i\}_{i=1}^n$ can be interpreted as the probability of pixel $i\in\{1,\dots,n\}$ not being changed.
The $\pi_i$ for the entire image are stored in a matrix $\boldsymbol{\Pi}\in[0,1]^{H\times W}$, from which the patch corresponding to $\boldsymbol{X}$ and $\boldsymbol{Y}$ is extracted and flattened into the vector $\boldsymbol{\pi}$.
These probabilities are not available at the beginning of training, so all entries of $\boldsymbol{\Pi}$ are initialised as $0$.
After several training epochs, a preliminary evaluation of the difference image $\boldsymbol{\Delta}$ is computed and scaled to fall into the range $[0,1]$, so that the prior can be updated as $\boldsymbol{\Pi} = 1 - \boldsymbol{\Delta}$.
In this way, pixels associated with a large $\boldsymbol{\Delta}$ entry are penalised by a small weight, whereas the opposite happens to pixels more likely to be unchanged.
The $\boldsymbol{\Pi}$ is updated iteratively at a rate that we can tune to accommodate both performance and computational cost.
This form of self-supervision paradigm has already proven robust in other tasks such as deep clustering~\cite{caron2018deep} and deep image recovery~\cite{ulyanov2018deep}.

The translation loss must be evaluated with paired data, since $\hat{\boldsymbol{X}}$ is computed from $\boldsymbol{Y}$ and compared with $\boldsymbol{X}$, while $\hat{\boldsymbol{Y}}$ is computed from $\boldsymbol{X}$ and compared with $\boldsymbol{Y}$. The code correlation loss, presented in the next section, also requires paired data.

\subsection{Code Correlation Loss}
The main contribution of this work lies in the way the codes are aligned. It therefore rests on the design and definition of the specific loss term associated with code alignment, referred to as the code correlation loss.

The distances in the input spaces between all pixel pairs $(i, j)$ in the co-located training patches are computed as $d^{\mathcal{X}}_{i,j}=d^{\mathcal{X}}(\boldsymbol{x}_{i},\boldsymbol{x}_{j})$ and $d^{\mathcal{Y}}_{i,j}=d^{\mathcal{Y}}(\boldsymbol{y}_{i},\boldsymbol{y}_{j})$ for $i,j\in\{1,\ldots,n\}$, where $\boldsymbol{x}_i$ and $\boldsymbol{y}_j$ denote the feature vectors of pixel $i\in\boldsymbol{X}$ and pixel $j\in\boldsymbol{Y}$, respectively.
The appropriate choice of distance measure depends on the underlying data distribution, but should also consider complexity.
The hypothesis of normality for imagery acquired by optical sensors is commonly assumed~\cite{bovolo2007theoretical,bovolo2015time}.
Concerning SAR intensity data, a logarithmic transformation is sufficient to bring it to near-Gaussianity~\cite{luppino2017clustering,zhan2018log}.
This qualifies the use of the computationally efficient Euclidean distance for both these data sources.

Once computed, the distances between all pixel pairs can be converted to the affinities
\begin{equation}
A^{\ell}_{i,j} =  \exp\left\{-\frac{\left(d^{\ell}_{i,j}\right)^2}{\sigma_{\ell}^2}\right\}\, \in \left(0,1\right], \quad i,j\in\{1,\dots,n\}\,.
\label{eq:affm}
\end{equation}
Here, $A^{\ell}_{i,j}$ are the entries of the affinity matrix $\boldsymbol{A}^{\ell} \in \mathbb{R}^{n \times n}$ for a given patch and modality $\ell\in\{\mathcal{X},\mathcal{Y}\}$, and
$\sigma_{\ell}$ is the kernel width, which must be automatically determined.
Our choice is to set it equal to the average distance to the $k^{th}$ nearest neighbour for all data points in the patch of modality $\ell$, with $k=\frac{3}{4}n$.
This heuristic, which can be traced back to \cite{mack1979multivariate}, captures the scale of local affinities within the patch and is robust with respect to outliers. 
Other common approaches to determine the kernel width, such as the Silverman's rule of thumb~\cite{wand1995kernel}, were discarded because they have not proven themselves as effective.

At this point, one can consider the rows
\begin{equation*}
A^{\mathcal{X}}_i = \big[A^{\mathcal{X}}_{i,1},\dots,A^{\mathcal{X}}_{i,n}\big]\;\text{and}\;
A^{\mathcal{Y}}_j = \big[A^{\mathcal{Y}}_{j,1},\dots,A^{\mathcal{Y}}_{j,n}\big]
\end{equation*}
as representations of pixel $i$ from patch $\boldsymbol{X}$ and pixel $j$ from patch $\boldsymbol{Y}$, respectively, in a new affinity space with $n$ features. Moreover, we can define a novel crossmodal distance between these pixels as
\begin{equation}
    D_{i,j} = \frac{1}{\sqrt{n}} \lVert A^{\mathcal{X}}_{i} - A^{\mathcal{Y}}_{j}\rVert_2\, \in \left[0,1\right], \; i,j\in\{1,\dots,n\},
\end{equation}
%
noting that since the affinities are normalised to the range $[0,1]$, then so is $D_{i,j}$.
This crossmodal distance allows to compare data across the two domains directly from their input space features. It further allows us to distinguish pixels that have consistent relations to other pixels in both domains from those that do not. This information can be interpreted in terms of probability of change.

The crossmodal input space distances $D_{i,j}$ for $i,j\in\{1,\dots,n\}$ are stored in $\boldsymbol{D}$. We next want to make sure that these are maintained in the code layer. We do this by defining similarities $S_{ij}=1-D_{ij}$ and forcing them to be as similar as possible to correlations between the codes of corresponding pixels. Let $\boldsymbol{z}^\mathcal{X}_i$ and $\boldsymbol{z}^\mathcal{Y}_j$ denote the entry of code patch $\boldsymbol{Z}^\mathcal{X}$ corresponding to pixel $i$ and the entry of code patch $\boldsymbol{Z}^\mathcal{Y}$ corresponding to pixel $j$, respectively. In mathematical terms, we enforce that
\begin{equation}
    R_{i,j}\triangleq\frac{\left(\boldsymbol{z}^{\mathcal{X}}_i\right)^T\boldsymbol{z}^{\mathcal{Y}}_j + |\mathcal{Z}|}{2\,|\mathcal{Z}|} \simeq S_{i,j}, \quad i,j\in\{1,\dots,n\}\,,
\label{eq:core}
\end{equation}
where the $S_{i,j}$ are elements of $\boldsymbol{S} = 1 - \boldsymbol{D}$. The normalisation of the codes, $\boldsymbol{z}^{\mathcal{X}}_i, \boldsymbol{z}^{\mathcal{Y}}_j \in [-1,1]^{|\mathcal{Z}|}$, and their dimensionality $|\mathcal{Z}|$ is such that the code correlations $R_{i,j}$ falls in the range $[0,1]$. Note that the elements on the diagonal of $\boldsymbol{S}$ represent the similarity between $\boldsymbol{x}_{i}$ and $\boldsymbol{y}_{i}$, that are not identical, so $S_{i,i}$ can be different from $1$.
Also observe that $\boldsymbol{S}$ is not symmetric, because the similarity between $\boldsymbol{x}_{i}$ and $\boldsymbol{y}_{j}$ is not necessarily the same as between $\boldsymbol{x}_{j}$ and $\boldsymbol{y}_{i}$.
%
%

Based on the above definitions and considerations, the code correlation loss term is defined as
\begin{equation}
    \mathcal{L}_{z} \left(\boldsymbol{\vartheta}\right) = \mathbb{E}_{\boldsymbol{X},\boldsymbol{Y}}\left[\delta(\boldsymbol{R},\boldsymbol{S})\right]\,,
\end{equation}
where the code correlation matrix $\boldsymbol{R}$ stores the $R_{i,j}$ from the left-hand side of Eq.\ \eqref{eq:core}.
Note that only encoder parameters are adjusted with this loss term.

\subsection{Total Loss Function}
Finally, the loss function minimised in this framework is the following weighted sum:
\begin{equation}
    \mathcal{L}\left(\boldsymbol{\vartheta}\right) = \lambda_{\mathrm r}\,\mathcal{L}_{\mathrm r}(\boldsymbol{\vartheta}) + \lambda_{\mathrm c}\,\mathcal{L}_{\mathrm c}(\boldsymbol{\vartheta})
    + \lambda_{\mathrm t}\,\mathcal{L}_{\mathrm t}(\boldsymbol{\vartheta}) + \lambda_{z}\,\mathcal{L}_{z} (\boldsymbol{\vartheta})\,,
\end{equation}
where the weights $\lambda_{\mathrm r}$, $\lambda_{\mathrm c}$, $\lambda_{\mathrm t}$ and $\lambda_{\mathrm z}$ are used to balance the loss terms and their impact on the optimisation result.
Together, the cycle-consistency and the code correlation let us achieve the sought alignment, while at the same time the other two terms keep focus on a correct reconstruction and transformation of the input.

After the training and the computation of $\boldsymbol{\Delta}$, the CD workflow includes an optional step and a mandatory step.
The former consists of spatial filtering of $\boldsymbol{\Delta}$ to reduce errors, based on the simple idea that spurious changed (unchanged) pixels surrounded by unchanged (changed) ones are most likely outliers that have been erroneously classified.
For our method we selected the Gaussian filtering presented in~\cite{krahenbuhl2011efficient}, which uses spatial context to regularise $\boldsymbol{\Delta}$.
The last step of a CD pipeline is to obtain the actual change map by thresholding $\boldsymbol{\Delta}$, and so all the pixels whose value is below the threshold are considered unchanged, vice versa for those with a larger value.
The optimal threshold can be found by visual inspection or automatically by exploiting an algorithm such as~\cite{kapur1985new,shanbhag1994utilization,yen1995new}.
We opted for the classical Otsu's method~\cite{otsu1979threshold}.

%% file: Results.tex
\section{Results}\label{sec:results}
\subsection{Implementation details}

%
%
For the proposed framework we deploy fully convolutional neural networks designed as follows:
$\text{Conv}(3\times 3 \times 100)\text{\textendash ReLU\textendash Conv}(3\times 3 \times 100)\text{\textendash ReLU\textendash Conv}(3\times 3 \times C)\text{\textendash Tanh}$.
$\text{Conv}(3\times 3 \times C)$ indicates a convolutional layer with $C$ filters of size $3 \times 3$, being $C = 3$ for the encoders, $C = |\mathcal{X}|$ for $D_{\mathcal{X}}$ and $C = |\mathcal{Y}|$ for $D_{\mathcal{Y}}$.
All the layers are non-strided and we apply padding to preserve the input size.
Leaky-ReLU~\cite{maas2013rectifier} with slope of $\beta=0.3$ for negative arguments is used.
Tanh indicates the hyperbolic tangent~\cite{maas2013rectifier}, which normalises data between $-1$ and $1$, as this has shown to speed up convergence~\cite{lecun2012efficient}.
Dropout~\cite{srivastava2014dropout} with a $20\%$ rate is applied.
A low number of features in the latent space allows to achieve the sought alignment more easily, whereas the number of layers and filters has been set to find a balance between flexibility of the network representations and the limited trainability of the networks, due to a small amount of training data.
Concerning the latter, at every epoch 10 batches containing 10 random patches of $100 \times 100$ pixels are extracted and randomly augmented ($90$ degrees rotations and upside-down flips).
As specified, the code correlation loss term $\mathcal{L}_{z}$ requires computation of a size $N\times N$ crossmodal distance matrix $\boldsymbol{D}$ when the training patch is $h\times w$. Due to memory constraints, only the inner $20 \times 20$ pixels of the training patches have been used to compute $\boldsymbol{D}$.
For normalisation of the matrix $\boldsymbol{D}$ between $0$ and $1$, the framework responded better when applying contrast stretching between the empirical batch minimum and maximum values of $\boldsymbol{D}$.
The four $\lambda$ values controlling the weighted sum of $\mathcal{L}$ were all set to $1$.

The Adam optimiser~\cite{reddi2018on} was selected to perform the minimisation of $\mathcal{L}$ for $100$ epochs with a learning rate of $10^{-4}$, which experienced a stair-cased exponential decay with rate $0.96$.
Actually, we found it beneficial to reduce the learning rate associated with $\mathcal{L}_z$ more aggressively with rate $0.9$.
This was implemented because it turned out most beneficial to correlate the code spaces at the beginning, when the autoencoder just started to learn a meaningful representation of the latent spaces and a reasonable transformation of the data. After some updates of $\boldsymbol{\Pi}$, $\mathcal{L}_z$ was experienced to function more as a regulariser, whereas the translation loss $\mathcal{L}_{\mathrm t}$ came more into play.
These updates were made every $25$ epochs, so at epoch $25$, $50$, and $75$.

\subsection{Evaluation criteria}

The performance of the proposed approach is measured in terms of two metrics.
The overall accuracy, $OA \in [0,1]$, is the ratio between correctly classified pixels and the total amount of pixels.
Cohen's kappa coefficient, $\kappa \in [-1,1]$, indicates the agreement between two classifiers~\cite{cohen1960coefficient}.
$\kappa = 1$ means total agreement, $\kappa = -1$ means total disagreement, $\kappa = 0$ means no correlation (random guess).
When comparing against a ground truth dataset, Cohen's kappa is expressed as
\begin{equation}
    \kappa = \frac{p_o - p_e}{1 - p_e}.
\end{equation}
Here, $p_o$ stands for the observed agreement between predictions and labels, i.e.\ the OA, while
$p_e$ is the probability of random agreement, which is estimated from the observed true positives (TP), true negatives (TN), false positives (FP), and false negatives (FN) as:
\begin{equation}
\begin{split}
    p_e = & \left(\frac{\text{TP} + \text{FP}}{N} \cdot \frac{\text{FN} + \text{TN}}{N}\right) \\  + & \left(\frac{\text{TP} + \text{FN}}{N} \cdot \frac{\text{FP} + \text{TN}}{N}\right)\,.
\end{split}
\end{equation}
In general, a high $\kappa$ implies a high $OA$, but not vice versa.
In any case, the papers presenting state-of-the-art methods do not always report both, so we compare algorithm performance dataset by dataset in terms of the available metrics.

\subsection{Methods compared}

 We will in the following present four datasets that are used to test the proposed method and reference algorithms. On the first two datasets, the
proposed method is compared to four similar deep learning approaches. The first two are the conditional adversarial network (CAN) of Niu et al.~\cite{niu2018conditional} and the symmetric convolutional coupling network (SCCN) of Liu et al.~\cite{liu2016deep}, which represent seminal work on unsupervised multimodal change detection with convolutional neural networks. The final two are are the ACE-Net and the X-Net recently proposed by the current authors in ~\cite{luppino2020deep}. To be aware of the characteristics of the training strategies employed by these methods, it should be noted that the CAN and the ACE-Net apply adversarial training, the ACE-Net and the SCCN exploit code alignment, while the ACE-Net and the X-Net use similar weighted image-to-image translation schemes as the proposed method.
The final two datasets have been used extensively by others in testing of methods whose source code we do not have access to. For these datasets we compare our results with the performance reported in Zhang et al. \cite{zhang2016cd} for post-classification comparison (PCC) and a deep learning model based on stacked denoising autoencoders (SDAE). We also compare with several methods proposed by Touati et al., namely a method that obtains its result my filtering a textural gradient-based similarity map (TGSM) \cite{touati2017new}, a method using energy-based encoding of nonlocal pairwise pixel interactions (EENPPI) \cite{touati2018energy}, a method based on modality invariant multidimensional scaling (MIMDS) \cite{touati2018change}, and a Markov model for multimodal change detection (M3CD) \cite{touati2019multimodal}. Finally, we compare with results obtained with the manifold learning-based statistical model (MLSM) of Prendes et al.~\cite{prendes2015new,prendes2015new2}.

\subsection{First dataset: Forest fire in Texas}

\input{Figures/Res/Texas/Input/Figures_input.tex}

A Landsat 5 Thematic Mapper (TM) multispectral image (Fig.\ \ref{fig:L5}) was acquired before a forest fire that took place in Bastrop County, Texas, during September-October 2011\footnote{Distributed by LP DAAC, http://lpdaac.usgs.gov \label{foot1}}.
An Earth Observing-1 Advanced Land Imager (EO-1 ALI) multispectral acquisition after the event completes the dataset (Fig. \ref{fig:ALI})$^1$.
Both images are optical, with $1534 \times 808$ pixels, and $7$ and $10$ channels respectively.
The ground truth of the event (see Fig.\ \ref{fig:gt1}) is provided by Volpi \textit{et al.}~\cite{volpi2015spectral}.

Fig.\ \ref{fig:kappa_Tx} displays the results obtained on this dataset by the proposed framework as compared to the reference methods. As one can notice, the proposed network produces consistently higher accuracy than the competitors and also maintains a low variance.
We also report that Volpi \textit{et al.}~\cite{volpi2015spectral} and Luppino \textit{et al.}~\cite{luppino2019unsupervised} achieved a $\kappa$ of $0.65$ and $0.91$ respectively with respect to the same ground truth.
Concerning the training times, their averages are listed in Table \ref{tab:times_Tx}.
These are comparable because the computation of the affinity matrices is time-consuming, but the proposed method is implemented with relatively small networks and trained for fewer iterations.

\begin{table}[ht!]
\centering
\caption{Average training time of the five methods on the Texas dataset.}
\label{tab:times_Tx}
\begin{tabular}{c | c | c | c | c}
\toprule
\small{CAN} & \small{SCCN} & \small{ACE-Net} & \small{X-Net} & \small{Proposed} \\
\midrule
$70$ min & $16$ min & $13$ min & $7$ min & $11$ min \\
\bottomrule
\end{tabular}
\end{table}

\input{Figures/Res/Texas/cohens_kappa.tex}

\subsection{Second dataset: Flood in California}

\input{Figures/Res/Cal/Input/Figures_input.tex}

Fig.\ \ref{fig:L8} shows the RGB channels of the Landsat 8 acquisition$^1$ covering Sacramento County, Yuba County and Sutter County, California, on 5 January 2017.
In addition, the multispectral sensors mounted on Landsat 8 provides another $8$ channels, going from deep blue to long-wave infrared.
The same area was affected by a flood, as it can be noticed in Fig.\ \ref{fig:S1A}.
This is a Sentinel-1A\footnote{Data processed by ESA, http://www.copernicus.eu/} acquisition, recorded in polarisations VV and VH on 18 February 2017 and augmented with the ratio between the two intensities as the third channel.
The ground truth in Fig.\ \ref{fig:cal_gt} is provided by Luppino \textit{et al.}~\cite{luppino2019unsupervised}.
Originally of $3500 \times 2000$ pixels, these images were resampled to $850 \times 500$ pixels as in~\cite{luppino2020deep} to compare the results. 

\begin{table}[ht!]
\centering
\caption{Average training time of the five methods on the California dataset.}
\label{tab:times_cal}
\begin{tabular}{c | c | c | c | c}
\toprule
\small{CAN} & \small{SCCN} & \small{ACE-Net} & \small{X-Net} & \small{Proposed} \\
\midrule
$21$ min & $15$ min & $12$ min & $6$ min & $8$ min \\
\bottomrule
\end{tabular}
\end{table}
\input{Figures/Res/Cal/cohens_kappa.tex}
The metrics obtained on this dataset are summarised in Fig.\ \ref{fig:kappa_cal}.
Also in this case, the proposed framework outperforms the state-of-the-art counterparts, both in terms of high quality and low variance.
For this dataset, $\kappa=0.46$ was achieved in~\cite{luppino2019unsupervised}.
Table \ref{tab:times_cal} contains the average training times on this dataset.
Again, the proposed approach required a training time which is in line with the state-of-the-art algorithms.

\subsection{Third dataset: Lake overflow in Italy}

\input{Figures/Res/Italy/Input/Figures_input.tex}

The next two datasets were provided by Touati \textit{et al.}~\cite{touati2019multimodal}.
In Fig.\ \ref{fig:it_NIR} and Fig.\ \ref{fig:it_RGB} are two Landsat 5 images of $412 \times 300$ pixels: the first is the Near InfraRed (NIR) band of an image acquired in September 1995, the second represents the red, green, and blue (RGB) bands sensed on the same area in July 1996.
These images were recorded before and after a lake overflow in Italy, whose profile is highlighted as ground truth in Fig.\ \ref{fig:it_gt}.
Table \ref{tab:acc_it} presents the average overall accuracy for several methods.
For the proposed method, the standard deviation is provided as well, and one may see that the results are very stable and close to the state-of-the-art.
The small amount of data in terms of the number of pixels does not in general favour deep learning approaches, and the relative performance could potentially change with larger training samples.
In this respect, Zhang \textit{et al.}~\cite{zhang2016cd} proposed a method that seems to be an exception, as this deep learning approach produces the best performance on this dataset.
However, it must be pointed out that, unlike us, they adapt their architectures to the dataset, which is infeasible in a completely unsupervised setting.
The average training time for the proposed framework on this dataset was a few seconds below $7$ minutes.

\begin{table}[t!]
\centering
\caption{Average accuracy of several methods on the lake overflow dataset. Best on top, proposed method in bold.}
\label{tab:acc_it}
\begin{tabular}{| c | c |}
\toprule
\textbf{Lake overflow dataset} & $OA$ \\
\midrule
SDAE~\cite{zhang2016cd} & $0.975$ \\
M3CD~\cite{touati2019multimodal} & $0.964$ \\
MIMDS~\cite{touati2018change} & $0.942$ \\
\textbf{Proposed} & $\boldsymbol{0.922 \pm 0.007}$ \\
PCC~\cite{zhang2016cd} & $0.882$ \\
\bottomrule
\end{tabular}
\end{table}

\subsection{Fourth dataset: Construction site in France}

\input{Figures/Res/France/Input/Figures_input.tex}

The last dataset includes two RGB images captured by Pleiades (Fig.\ \ref{fig:pleiades}) and WorldView 2 (Fig.\ \ref{fig:Wview}), showing the work progress of road constructions in Toulouse, France, during May 2012 and July 2013.
The ground truth in Fig.\ \ref{fig:fr_gt} depicts such progress.
For computational reasons, the images were reduced from $2000 \times 2000$ pixels to $500 \times 500$ as in~\cite{touati2019multimodal}, leading to an average training time of $7$ minutes.
The average accuracy obtained by several methods on this dataset is listed in Table \ref{tab:acc_fr}.
Again, the accuracy of the proposed method comes with a standard deviation, and also in this case it is very stable and close to the state-of-the-art.

\begin{table}[b!]
\centering
\caption{Average accuracy of several methods on the constructions dataset. Best on top, proposed method in bold.}
\label{tab:acc_fr}
\begin{tabular}{| c | c |}
\toprule
\textbf{Constructions dataset} & $OA$ \\
\midrule
MIMDS~\cite{touati2018change} & $0.877$ \\
TGSM~\cite{touati2017new} & $0.870$ \\
M3CD~\cite{touati2019multimodal} & $0.862$ \\
\textbf{Proposed} & $\boldsymbol{0.859 \pm 0.003}$ \\
EENPPI~\cite{touati2018energy} & $0.853$ \\
MLSM~\cite{prendes2015new} & $0.844$ \\
\bottomrule
\end{tabular}
\end{table}

\input{Figures/Res/Examples.tex}

Finally, in Fig.\ \ref{fig:examples} we present a visual example of the transformations obtained with the proposed method on the datasets used in this section.
As it can be seen, the data from one input domain are transformed into the other in a meaningful way, and the resemblance between the styles of the fake images and the original images is clear.
In the two last datasets, one could speculate that the low amount of data and features (few pixels consisting of few channels) did not allow to achieve a proper alignments of the code spaces.
This endorses the choice to compute $\boldsymbol{d}$ as a weighted sum of the difference images in the input spaces rather than just the difference image in the latent space, although it still remains a valid option.

%% file: Figures/Res/Texas/Input/Figures_input.tex
\begin{figure}[h!]

\begin{subfigure}[t]{0.3\columnwidth}
\includegraphics[width=\linewidth,keepaspectratio]{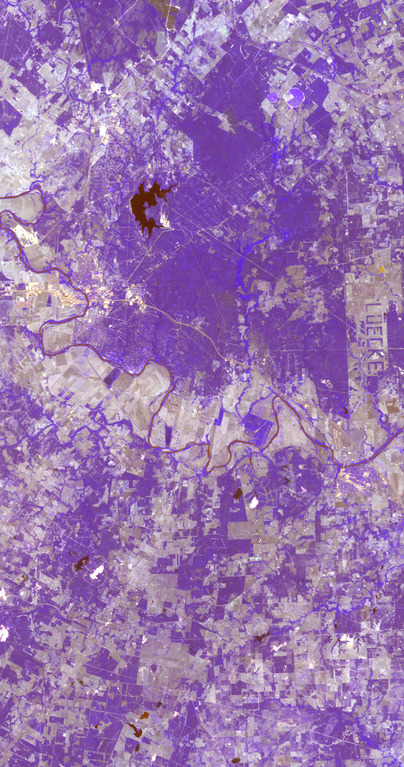}
\caption{Landsat 5 ($t_1$)}
\label{fig:L5}
\end{subfigure}
\hfill
\begin{subfigure}[t]{0.3\columnwidth}
\includegraphics[width=\linewidth,keepaspectratio]{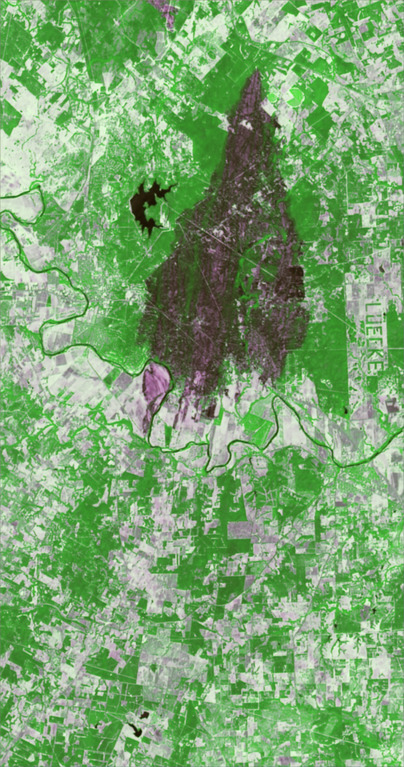}
\caption{EO-1 ALI ($t_2$)}
\label{fig:ALI}
\end{subfigure}
\hfill
\begin{subfigure}[t]{0.3\columnwidth}
\includegraphics[width=\linewidth,keepaspectratio]{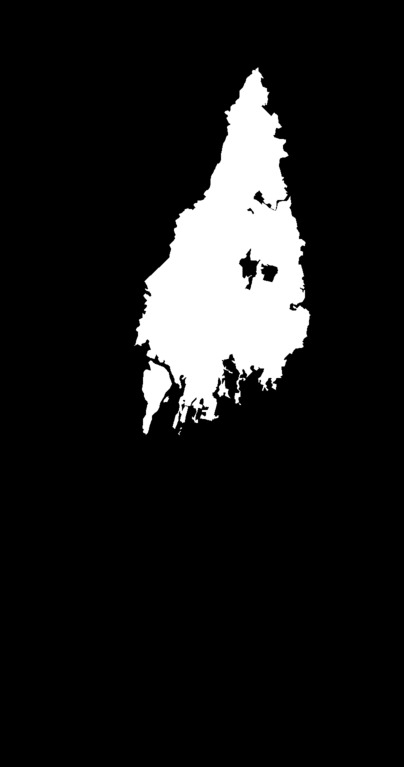}
\caption{Ground Truth}
\label{fig:gt1}
\end{subfigure}
\caption{Forest fire in Texas: Landsat 5 ($t1$), (b) EO-1 ALI ($t2$), (c) ground truth. RGB false color composites are shown for both images.}
\label{fig:dataset1}
\end{figure}

%% file: Figures/Res/Texas/cohens_kappa.tex
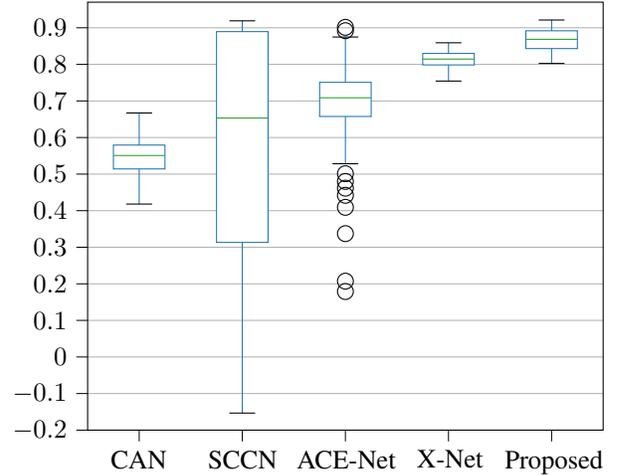
\begin{figure}[h]
\begin{tikzpicture}

\definecolor{color0}{rgb}{0.12156862745098,0.466666666666667,0.705882352941177}
\definecolor{color1}{rgb}{0.172549019607843,0.627450980392157,0.172549019607843}

\begin{axis}[
tick align=outside,
tick pos=left,
x grid style={white!69.01960784313725!black},
xmin=0.5, xmax=5.5,
xtick style={color=black},
xtick={3,1,5,2,4},
xticklabels={,,,,},
y grid style={white!69.01960784313725!black},
ymajorgrids=true,
major grid style = {line width = 0.1pt, draw = gray!},
ytick distance=0.1,
ymin=-0.2, ymax=0.97,
ytick style={color=black}
]

\filldraw[fill=white, draw=color0] (2.75, 0.65755686925) rectangle (3.25, 0.7508778645);
\addplot [color0, opacity=1]
table {%
3 0.65755686925
3 0.528298141
};
\addplot [color0, opacity=1]
table {%
3 0.7508778645
3 0.874739937
};
\addplot [black]
table {%
2.875 0.528298141
3.125 0.528298141
};
\addplot [black]
table {%
2.875 0.874739937
3.125 0.874739937
};
\addplot [black, mark=*, mark size=3, mark options={solid,fill opacity=0}, only marks]
table {%
3 0.3372163
3 0.501075035
3 0.44200156
3 0.461528641
3 0.207245703
3 0.47995946
3 0.409150494
3 0.178960661
3 0.891986466
3 0.901247412
};
\filldraw[fill=white, draw=color0] (0.75, 0.51412788175) rectangle (1.25, 0.579346398);
\addplot [color0, opacity=1]
table {%
1 0.51412788175
1 0.418049666
};
\addplot [color0, opacity=1]
table {%
1 0.579346398
1 0.667111636
};
\addplot [black]
table {%
0.875 0.418049666
1.125 0.418049666
};
\addplot [black]
table {%
0.875 0.667111636
1.125 0.667111636
};
\filldraw[fill=white, draw=color0] (4.75, 0.8431773785) rectangle (5.25, 0.891758382);
\addplot [color0, opacity=1]
table {%
5 0.8431773785
5 0.802601755
};
\addplot [color0, opacity=1]
table {%
5 0.891758382
5 0.921224296
};
\addplot [black]
table {%
4.875 0.802601755
5.125 0.802601755
};
\addplot [black]
table {%
4.875 0.921224296
5.125 0.921224296
};
\filldraw[fill=white, draw=color0] (1.75, 0.31325439675) rectangle (2.25, 0.8895221615);
\addplot [color0, opacity=1]
table {%
2 0.31325439675
2 -0.153350765
};
\addplot [color0, opacity=1]
table {%
2 0.8895221615
2 0.919036682
};
\addplot [black]
table {%
1.875 -0.153350765
2.125 -0.153350765
};
\addplot [black]
table {%
1.875 0.919036682
2.125 0.919036682
};
\filldraw[fill=white, draw=color0] (3.75, 0.798548381) rectangle (4.25, 0.829944861);
\addplot [color0, opacity=1]
table {%
4 0.798548381
4 0.754368963
};
\addplot [color0, opacity=1]
table {%
4 0.829944861
4 0.859014775
};
\addplot [black]
table {%
3.875 0.754368963
4.125 0.754368963
};
\addplot [black]
table {%
3.875 0.859014775
4.125 0.859014775
};
\addplot [color1, opacity=1]
table {%
2.75 0.708680162
3.25 0.708680162
};
\addplot [color1, opacity=1]
table {%
0.75 0.5508208445
1.25 0.5508208445
};
\addplot [color1, opacity=1]
table {%
4.75 0.868460715
5.25 0.868460715
};
\addplot [color1, opacity=1]
table {%
1.75 0.653262889
2.25 0.653262889
};
\addplot [color1, opacity=1]
table {%
3.75 0.8144591625
4.25 0.8144591625
};
\end{axis}
\node[align=center, below] at (0.68,-0.143)%
{CAN};
\node[align=center, below] at (2.06,-0.143)%
{SCCN};
\node[align=center, below] at (3.44,-0.143)%
{ACE-Net};
\node[align=center, below] at (4.82,-0.143)%
{X-Net};
\node[align=center, below] at (6.2,-0.143)%
{Proposed};

\end{tikzpicture}
\caption{$\kappa$ obtained on the Texas dataset by the proposed approach and several state-of-the-art methods.}
\label{fig:kappa_Tx}
\end{figure}

%% file: Figures/Res/Cal/Input/Figures_input.tex
\begin{figure}[ht!]

\begin{subfigure}[t]{0.3\columnwidth}
\includegraphics[width=\linewidth,keepaspectratio]{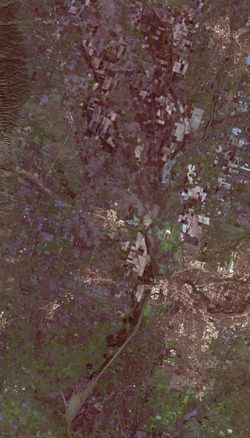}
\caption{Landsat 8 ($t_1$)}
\label{fig:L8}
\end{subfigure}
\hfill
\begin{subfigure}[t]{0.3\columnwidth}
\includegraphics[width=\linewidth,keepaspectratio]{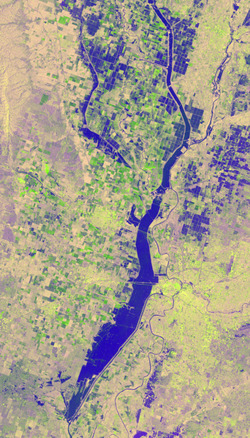}
\caption{Sentinel1-A ($t_2$)}
\label{fig:S1A}
\end{subfigure}
\hfill
\begin{subfigure}[t]{0.3\columnwidth}
\includegraphics[width=\linewidth,keepaspectratio]{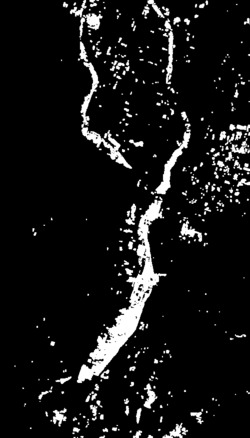}
\caption{Ground Truth}
\label{fig:cal_gt}
\end{subfigure}
\caption{Flood in California: Landsat 8 ($t1$), (b) Sentinel1-A ($t2$), (c) ground truth. RGB false color composites are shown for both images.}
\label{fig:california}
\end{figure}

%% file: Figures/Res/Cal/cohens_kappa.tex
\begin{figure}[ht!]
    \centering

\begin{tikzpicture}

\definecolor{color0}{rgb}{0.12156862745098,0.466666666666667,0.705882352941177}
\definecolor{color1}{rgb}{0.172549019607843,0.627450980392157,0.172549019607843}

\begin{axis}[
tick align=outside,
tick pos=left,
x grid style={white!69.01960784313725!black},
xmin=0.5, xmax=5.5,
xtick style={color=black},
xtick={3,1,5,2,4},
xticklabels={,,,,},
y grid style={white!69.01960784313725!black},
ymajorgrids=true,
major grid style = {line width = 0.1pt, draw = gray!},
ytick distance=0.05,
ymin=0.2, ymax=0.53,
ytick style={color=black}
]
\filldraw[fill=white, draw=color0] (2.75, 0.3811055715) rectangle (3.25, 0.43223964225);
\addplot [color0, opacity=1]
table {%
3 0.3811055715
3 0.314152558
};
\addplot [color0, opacity=1]
table {%
3 0.43223964225
3 0.477597232
};
\addplot [black]
table {%
2.875 0.314152558
3.125 0.314152558
};
\addplot [black]
table {%
2.875 0.477597232
3.125 0.477597232
};
\addplot [black, mark=*, mark size=3, mark options={solid,fill opacity=0}, only marks]
table {%
3 0.258354047
3 0.223756168
3 0.276487195
3 0.293815267
3 0.295611357
3 0.24720199
};
\filldraw[fill=white, draw=color0] (0.75, 0.34852941775) rectangle (1.25, 0.3788557725);
\addplot [color0, opacity=1]
table {%
1 0.34852941775
1 0.315602196
};
\addplot [color0, opacity=1]
table {%
1 0.3788557725
1 0.42038473
};
\addplot [black]
table {%
0.875 0.315602196
1.125 0.315602196
};
\addplot [black]
table {%
0.875 0.42038473
1.125 0.42038473
};
\filldraw[fill=white, draw=color0] (4.75, 0.4610895965) rectangle (5.25, 0.48932907);
\addplot [color0, opacity=1]
table {%
5 0.4610895965
5 0.435423374
};
\addplot [color0, opacity=1]
table {%
5 0.48932907
5 0.52270931
};
\addplot [black]
table {%
4.875 0.435423374
5.125 0.435423374
};
\addplot [black]
table {%
4.875 0.52270931
5.125 0.52270931
};
\filldraw[fill=white, draw=color0] (1.75, 0.44313298425) rectangle (2.25, 0.46552772175);
\addplot [color0, opacity=1]
table {%
2 0.44313298425
2 0.422397597
};
\addplot [color0, opacity=1]
table {%
2 0.46552772175
2 0.483213411
};
\addplot [black]
table {%
1.875 0.422397597
2.125 0.422397597
};
\addplot [black]
table {%
1.875 0.483213411
2.125 0.483213411
};
\addplot [black, mark=*, mark size=3, mark options={solid,fill opacity=0}, only marks]
table {%
2 0.394507441
2 0.399173554
2 0.408103218
};
\filldraw[fill=white, draw=color0] (3.75, 0.3721866195) rectangle (4.25, 0.40495477675);
\addplot [color0, opacity=1]
table {%
4 0.3721866195
4 0.33419473
};
\addplot [color0, opacity=1]
table {%
4 0.40495477675
4 0.444436246
};
\addplot [black]
table {%
3.875 0.33419473
4.125 0.33419473
};
\addplot [black]
table {%
3.875 0.444436246
4.125 0.444436246
};
\addplot [black, mark=*, mark size=3, mark options={solid,fill opacity=0}, only marks]
table {%
4 0.312105881
};
\addplot [color1, opacity=1]
table {%
2.75 0.4060308935
3.25 0.4060308935
};
\addplot [color1, opacity=1]
table {%
0.75 0.365811749
1.25 0.365811749
};
\addplot [color1, opacity=1]
table {%
4.75 0.4799793365
5.25 0.4799793365
};
\addplot [color1, opacity=1]
table {%
1.75 0.457497096
2.25 0.457497096
};
\addplot [color1, opacity=1]
table {%
3.75 0.3891022685
4.25 0.3891022685
};
\end{axis}
\node[align=center, below] at (0.68,-0.143)%
{CAN};
\node[align=center, below] at (2.06,-0.143)%
{SCCN};
\node[align=center, below] at (3.44,-0.143)%
{ACE-Net};
\node[align=center, below] at (4.82,-0.143)%
{X-Net};
\node[align=center, below] at (6.2,-0.143)%
{Proposed};

\end{tikzpicture}
\caption{$\kappa$ obtained on the California dataset by the proposed approach and several state-of-the-art methods}
    \label{fig:kappa_cal}
\end{figure}
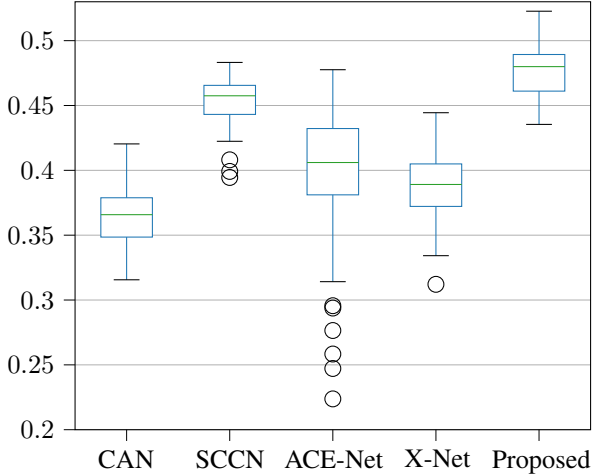

%% file: Figures/Res/Italy/Input/Figures_input.tex
\begin{figure}[ht!]

\begin{subfigure}[t]{0.3\columnwidth}
\includegraphics[width=\linewidth,keepaspectratio]{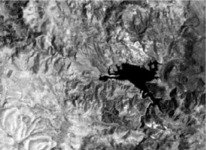}
\caption{Landsat 5 NIR($t_1$)}
\label{fig:it_NIR}
\end{subfigure}
\hfill
\begin{subfigure}[t]{0.3\columnwidth}
\includegraphics[width=\linewidth,keepaspectratio]{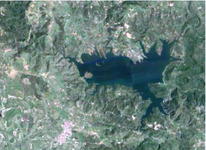}
\caption{Landsat 5 RGB ($t_2$)}
\label{fig:it_RGB}
\end{subfigure}
\hfill
\begin{subfigure}[t]{0.3\columnwidth}
\includegraphics[width=\linewidth,keepaspectratio]{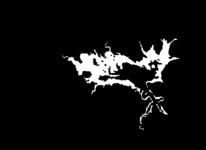}
\caption{Ground Truth}
\label{fig:it_gt}
\end{subfigure}
\caption{Lake overflow in Italy: Landsat 5 Near InfraRed (NIR) band ($t1$), (b) Landsat 5 red, green, and blue (RGB) bands ($t2$), (c) ground truth.}
\label{fig:italy}
\end{figure}

%% file: Figures/Res/France/Input/Figures_input.tex
\begin{figure}[h!]

\begin{subfigure}[t]{0.3\columnwidth}
\includegraphics[width=\linewidth,keepaspectratio]{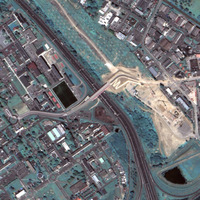}
\caption{Pleiades ($t_1$)}
\label{fig:pleiades}
\end{subfigure}
\hfill
\begin{subfigure}[t]{0.3\columnwidth}
\includegraphics[width=\linewidth,keepaspectratio]{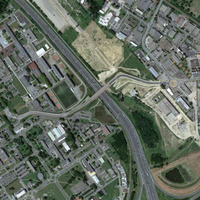}
\caption{WorldView 2 ($t_2$)}
\label{fig:Wview}
\end{subfigure}
\hfill
\begin{subfigure}[t]{0.3\columnwidth}
\includegraphics[width=\linewidth,keepaspectratio]{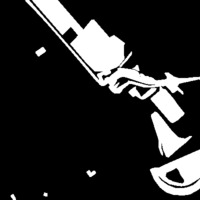}
\caption{Ground Truth}
\label{fig:fr_gt}
\end{subfigure}
\caption{Constructions in France: Pleiades ($t1$), (b) WorldView 2 ($t2$), (c) ground truth.}
\label{fig:france}
\end{figure}

%% file: Figures/Res/Examples.tex
\newlength{\tempdima}
\newlength{\tempdimb}
\newlength{\tempdimc}
\newlength{\tempdimd}
\newcommand{\rownamea}[1]
{\rotatebox{90}{\makebox[\tempdima][c]{\textbf{#1}}}}
\newcommand{\rownameb}[1]
{\rotatebox{90}{\makebox[\tempdimb][c]{\textbf{#1}}}}
\newcommand{\rownamec}[1]
{\rotatebox{90}{\makebox[\tempdimc][c]{\textbf{#1}}}}
\newcommand{\rownamed}[1]
{\rotatebox{90}{\makebox[\tempdimd][c]{\textbf{#1}}}}

\begin{figure*}
\settoheight{\tempdima}{\includegraphics[width=.11\linewidth]{Figures/Res/Texas/Input/X.jpg}}%
\settoheight{\tempdimb}{\includegraphics[width=.11\linewidth]{Figures/Res/Cal/Input/X.jpg}}%
\settoheight{\tempdimc}{\includegraphics[width=.11\linewidth]{Figures/Res/Italy/Input/X.jpg}}%
\settoheight{\tempdimd}{\includegraphics[width=.11\linewidth]{Figures/Res/France/Input/X.jpg}}%
\centering\begin{tabular}{@{}c@{ }c@{ }c@{ }c@{ }c@{ }c@{ }c@{ }c@{ }c@{}}
\rownamea{Texas}&
\includegraphics[width=.11\linewidth]{Figures/Res/Texas/Input/X.jpg}&
\includegraphics[width=.11\linewidth]{Figures/Res/Texas/Input/Y.jpg}&
\includegraphics[width=.11\linewidth]{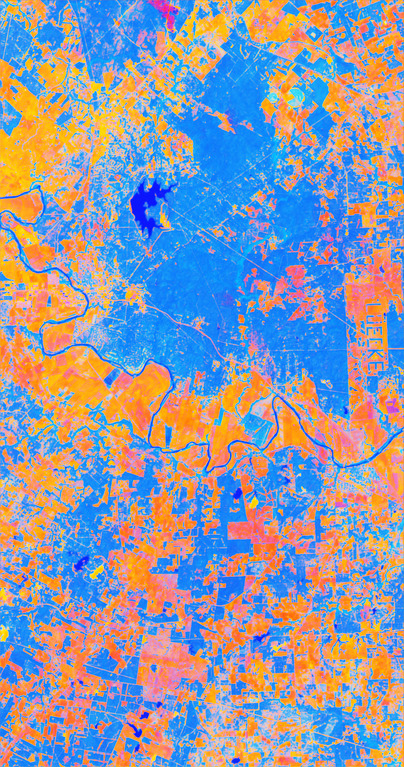}&
\includegraphics[width=.11\linewidth]{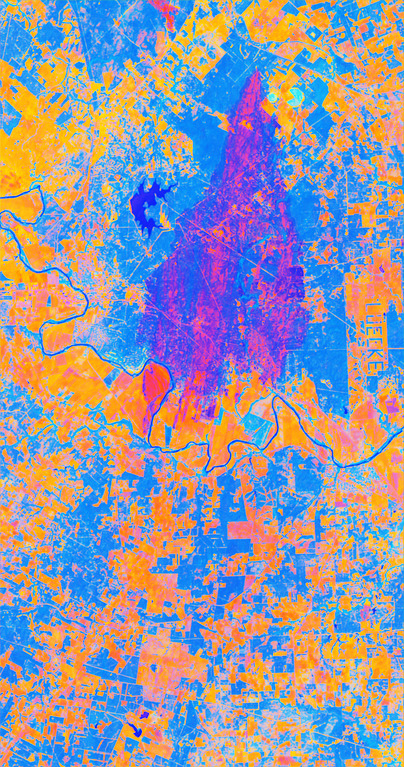}&
\includegraphics[width=.11\linewidth]{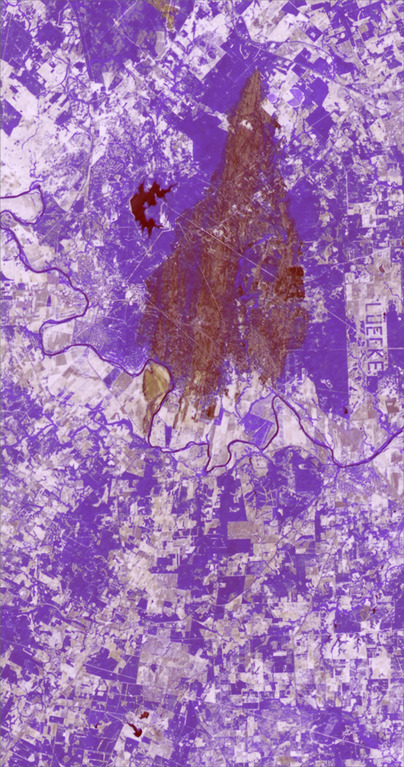}&
\includegraphics[width=.11\linewidth]{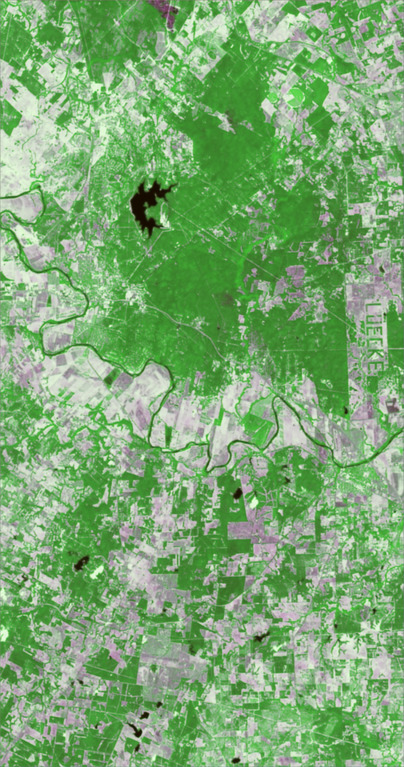}&
\includegraphics[width=.11\linewidth]{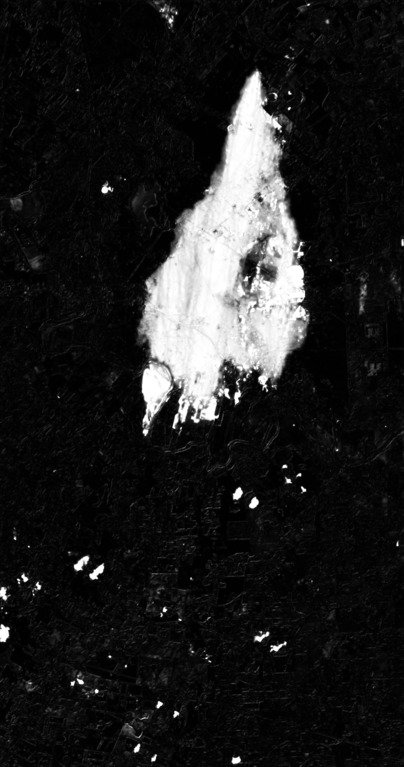}&
\includegraphics[width=.11\linewidth]{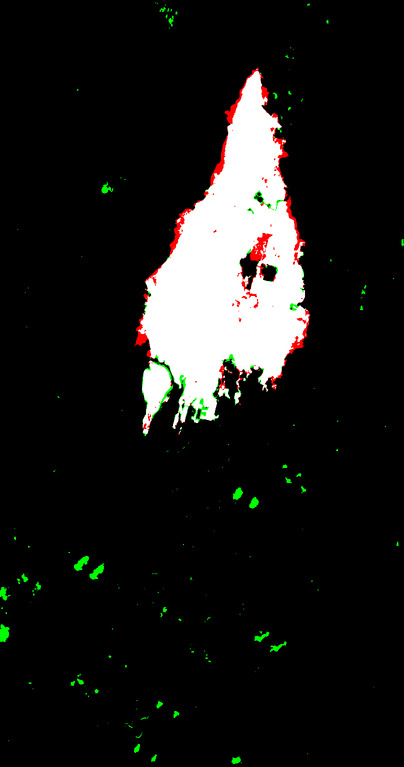}\\
&$\boldsymbol{X}$ & $\boldsymbol{Y}$ & $\boldsymbol{Z}^\mathcal{X}$ & $\boldsymbol{Z}^\mathcal{Y}$ & $\boldsymbol{\hat{X}}$ & $\boldsymbol{\hat{Y}}$ & $\boldsymbol{d}$ filtered & Confusion map \\
\rownameb{California}&
\includegraphics[width=.11\linewidth]{Figures/Res/Cal/Input/X.jpg}&
\includegraphics[width=.11\linewidth]{Figures/Res/Cal/Input/Y.jpg}&
\includegraphics[width=.11\linewidth]{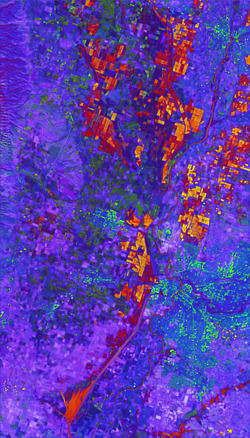}&
\includegraphics[width=.11\linewidth]{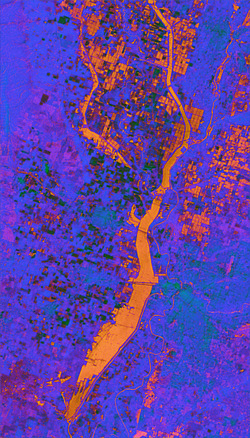}&
\includegraphics[width=.11\linewidth]{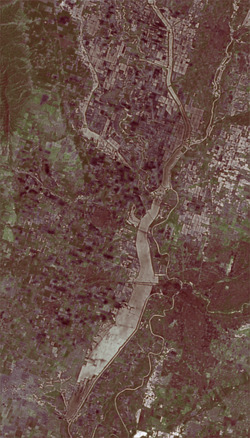}&
\includegraphics[width=.11\linewidth]{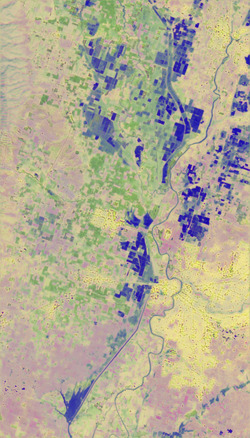}&
\includegraphics[width=.11\linewidth]{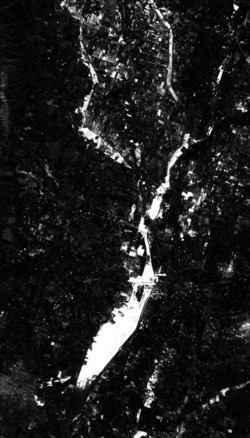}&
\includegraphics[width=.11\linewidth]{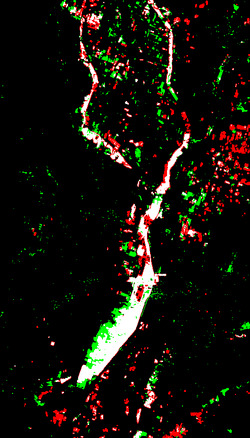}\\
&$\boldsymbol{X}$ & $\boldsymbol{Y}$ & $\boldsymbol{Z}^\mathcal{X}$ & $\boldsymbol{Z}^\mathcal{Y}$ & $\boldsymbol{\hat{X}}$ & $\boldsymbol{\hat{Y}}$ & $\boldsymbol{d}$ filtered & Confusion map \\
\rownamec{Italy}&
\includegraphics[width=.11\linewidth]{Figures/Res/Italy/Input/X.jpg}&
\includegraphics[width=.11\linewidth]{Figures/Res/Italy/Input/Y.jpg}&
\includegraphics[width=.11\linewidth]{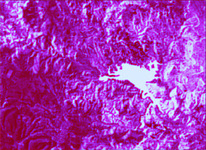}&
\includegraphics[width=.11\linewidth]{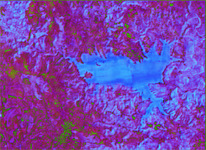}&
\includegraphics[width=.11\linewidth]{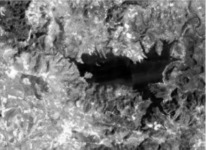}&
\includegraphics[width=.11\linewidth]{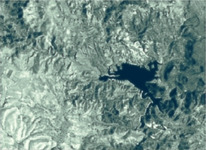}&
\includegraphics[width=.11\linewidth]{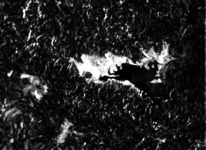}&
\includegraphics[width=.11\linewidth]{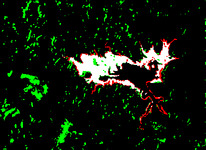}\\
&$\boldsymbol{X}$ & $\boldsymbol{Y}$ & $\boldsymbol{Z}^\mathcal{X}$ & $\boldsymbol{Z}^\mathcal{Y}$ & $\boldsymbol{\hat{X}}$ & $\boldsymbol{\hat{Y}}$ & $\boldsymbol{d}$ filtered & Confusion map \\
\rownamed{France}&
\includegraphics[width=.11\linewidth]{Figures/Res/France/Input/X.jpg}&
\includegraphics[width=.11\linewidth]{Figures/Res/France/Input/Y.jpg}&
\includegraphics[width=.11\linewidth]{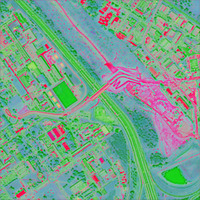}&
\includegraphics[width=.11\linewidth]{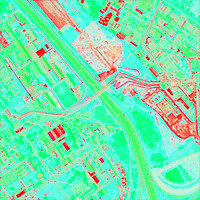}&
\includegraphics[width=.11\linewidth]{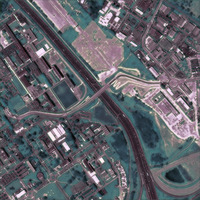}&
\includegraphics[width=.11\linewidth]{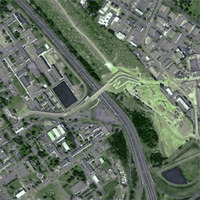}&
\includegraphics[width=.11\linewidth]{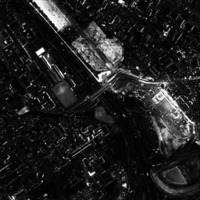}&
\includegraphics[width=.11\linewidth]{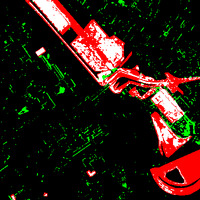}\\
&$\boldsymbol{X}$ & $\boldsymbol{Y}$ & $\boldsymbol{Z}^\mathcal{X}$ & $\boldsymbol{Z}^\mathcal{Y}$ & $\boldsymbol{\hat{X}}$ & $\boldsymbol{\hat{Y}}$ & $\boldsymbol{d}$ filtered & Confusion map \\
\end{tabular}
\caption{Examples of final results, organized in one row for each dataset. Col.\ $1$: input image $\boldsymbol{X}$; Col.\ $2$: input image $\boldsymbol{Y}$; Col.\ $3$: transformations of $\boldsymbol{X}$ into the code space $\boldsymbol{Z}^\mathcal{X} = E_\mathcal{X}(\boldsymbol{X})$; Col.\ $4$: transformations of $\boldsymbol{Y}$ into the code space $\boldsymbol{Z}^\mathcal{Y} = E_\mathcal{Y}(\boldsymbol{Y})$; Col.\ $5$ transformations $\boldsymbol{\hat{Y}} = F(\boldsymbol{X})$; Col.\ $6$: transformations $\boldsymbol{\hat{X}} = G(\boldsymbol{Y})$; Col.\ $7$: $\boldsymbol{d}$ filtered; Col.\ $8$: Confusion map (TP: white; TN: black; FP: green; FN: red) (g)}
\label{fig:examples}
\end{figure*}

%% file: Conclusions.tex
\section{Conclusions}\label{sec:conclusions}

In this work, we presented a novel unsupervised methodology to align the code spaces of two autoencoders based on affinity information extracted from the input data.
In particular, this is part of a heterogeneous CD framework that allows to achieve this latent space entanglement even when the input images contain changes, whose misleading contribution to the training is considerably reduced.
The method proved to perform consistently on par with or better than the state-of-the-art across four different datasets.
Its performance worsen when handling a limited amount of features in input, especially when only one channel is available in one of the images, implying a regression from one variable to many, which is an ill-posed problem.
On the other hand, it deals properly with multispectral and multipolarisation images, by being able to map data appropriately across domains in a meaningful manner.

\section{Acknowledgement}

The project and the first author was funded by the Research Council of Norway under research grant no.\ 251327. We gratefully acknowledge the support of NVIDIA Corporation by the donation of the GPU used for this research.